\pdfoutput=1

\documentclass[11pt]{article}

\usepackage[final]{acl}

\usepackage{times}
\usepackage{latexsym}

\usepackage[T1]{fontenc}

\usepackage[utf8]{inputenc}

\usepackage{microtype}

\usepackage[export]{adjustbox}

\usepackage{booktabs}
\usepackage{physics}
\usepackage{amsmath}
\usepackage{tikz}
\usepackage{mathdots}
\usepackage{yhmath}
\usepackage{cancel}
\usepackage{color}
\usepackage{siunitx}
\usepackage{array}
\usepackage{multirow}
\usepackage{amssymb}
\usepackage{gensymb}
\usepackage{tabularx}
\usepackage{extarrows}
\usepackage{booktabs}
\usepackage{caption}
\usepackage{subcaption}
\usepackage{pifont}
\usepackage{xcolor}
\usepackage{graphicx}
\usetikzlibrary{fadings}
\usetikzlibrary{patterns}
\usetikzlibrary{shadows.blur}
\usetikzlibrary{shapes}

\def \t{{\bf t}}
\def \e{{\bf e}}
\def \v{{\bf v}}

\def \w{{\bf w}}
\def \x{{\bf x}}
\def \W{{\bf W}}
\def \I{{\bf I}}
\def \X{{\bf X}}
\def \V{{\bf V}}
\def \E{{\bf E}}
\def \C{{\bf C}}
\def \T{{\bf T}}

\definecolor{tablegray}{RGB}{117,117,117}

\title{Enabling Multimodal Generation on CLIP via Vision-Language \\Knowledge Distillation}

\author{Wenliang Dai$^1$, Lu Hou$^2$, Lifeng Shang$^2$, Xin Jiang$^2$, Qun Liu$^2$, Pascale Fung$^1$ \\
$^1$Hong Kong University of Science and Technology, 
$^2$Huawei Noah’s Ark Lab \\
wdaiai@connect.ust.hk, pascale@ece.ust.hk, \\
\{houlu3, shang.lifeng, jiang.xin, qun.liu\}@huawei.com}

\begin{document}
\maketitle
\begin{abstract}
The recent large-scale vision-language pre-training (VLP) of dual-stream architectures (e.g., CLIP) with a tremendous amount of image-text pair data, has shown its superiority on various multimodal alignment tasks.
Despite its success, the resulting models are not capable of multimodal generative tasks due to the weak text encoder. 
To tackle this problem, we propose to augment the dual-stream VLP model with a textual pre-trained language model (PLM) via vision-language knowledge distillation (VLKD), enabling the capability for multimodal generation. VLKD is pretty data- and computation-efficient compared to the pre-training from scratch. 
Experimental results show that the resulting model 
has 
strong zero-shot performance on multimodal generation tasks, such as open-ended visual question answering and image captioning. 
For example, it achieves 44.5\% zero-shot accuracy on the VQAv2 dataset, surpassing the previous state-of-the-art zero-shot model with $7\times$ fewer parameters. Furthermore, the original textual language understanding and generation ability of the PLM is maintained after VLKD, which makes our model versatile for both multimodal and unimodal tasks.
\end{abstract}

\begin{figure}[t]
    \centering
    \includegraphics[width=\linewidth]{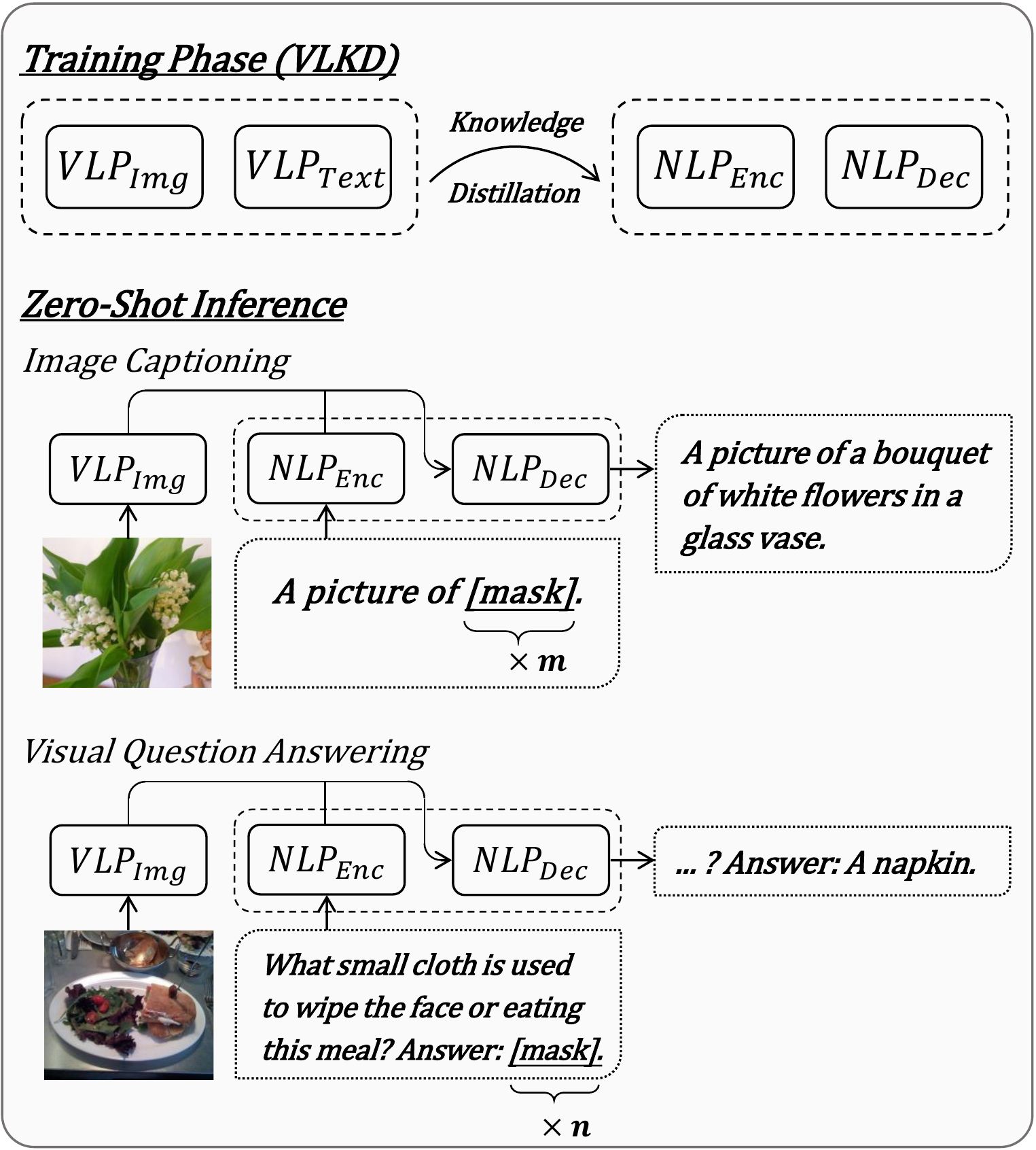}
    \caption{Intuition of our proposed approach. After VLKD, the model can fill in the masked locations with meaningful words to describe the image without further finetuning. Moreover, it can answer questions with proper reasoning over the given images and pre-trained knowledge inside PLMs, e.g., \textit{a napkin} is for wiping the face at meals.}
    \label{fig:intro}
\end{figure}

\section{Introduction} \label{sec:introduction}
Recent large-scale dual-stream Vision-Language Pre-training (VLP) models like CLIP~\citep{clip} and ALIGN~\citep{align}, have shown 
remarkable performance on various downstream multimodal alignment tasks, e.g.,  image-text retrieval and image classification. 
These models are pre-trained using cross-modal contrastive learning on 
tremendous
image-text pairs and learn strong multimodal 
representations.
Despite their success, as mentioned by \citet{clip}, their text encoder is relatively weak by only having a discriminative multimodal
pre-training objective, which makes them incompetent on generative multimodal tasks such as image captioning and open-ended visual question answering (VQA).

Meanwhile, the Transformer-based~\citep{transformer} auto-regressive large-scale pre-trained language models (PLMs), such as GPT~\citep{gpt,gpt3}, have been dominating in the natural language generation (NLG) tasks.
These models are usually trained with causal self-attention, which only allows the model to attend to past outputs (unidirectional) to satisfy their generative nature.
More recently, BART~\cite{bart} and T5~\cite{t5} propose to augment the auto-regressive decoder with a bidirectional Transformer encoder
to further capture bidirectional information of the input.
These encoder-decoder 
architectures excel on not only NLG but also understanding (NLU) tasks.


To tackle the aforementioned limitations of dual-stream VLP models and 
fully utilize
PLMs, in this paper, we present \textit{\textbf{V}}ision-\textit{\textbf{L}}anguage \textit{\textbf{K}}nowledge \textit{\textbf{D}}istillation (\textit{\textbf{VLKD}}), a simple yet effective approach to enable CLIP
to perform generative multimodal tasks through knowledge distillation.
Specifically, we align the BART encoder to CLIP's joint multimodal embedding space to gain the understanding of multimodal knowledge, along with
an image-conditioned language modeling loss to consort BART encoder and decoder. 
During training, we freeze CLIP's weights to keep its learned
multimodal space. 
For the finetuning and inference of downstream tasks, the original CLIP text encoder is discarded, which can be interpreted as being replaced by the distilled BART.
Therefore, we leverage the strengths from both sides, the expressive multimodal representation space of CLIP and the strong text generation capability of BART.

Compared to VLP from scratch, VLKD uses several magnitudes fewer image-text pairs 
and computational resources. As depicted in Figure~\ref{fig:intro}, after VLKD pre-training, the model exhibits strong zero-shot performance on generative multimodal tasks, including open-ended VQA and image captioning. 
Without finetuning, it has the ability to generate answers by reasoning over the question, the visual information, and the textual knowledge embedded in the pre-trained BART. Furthermore, it can also directly generate a plausible caption given an image. Empirical results show that our model achieves 44.5\% accuracy on the VQAv2 dataset and 84.6 CIDEr on COCO image caption dataset in a zero-shot manner. 
Moreover, the original NLU and NLG ability of BART is maintained, which makes the model versatile for both multimodal and unimodal tasks. 

To summarize, our contributions are: 1) We introduce an 
efficient
approach to distill knowledge from the 
dual-stream VLP model CLIP to BART.
The resulting
model shows strong zero-shot performance on generative multimodal tasks, as well as pure NLP tasks; 2) We exhaustively quantify these capabilities on six benchmarks under various settings; and 3) We conduct comprehensive analysis and ablation study to provide insights and grease future work on this direction.

\section{Related Work} \label{sec:related_work}
\subsection{Vision-language Pre-training}
Based on how the two modalities interact, recent VLP models mainly fall into two categories: single-stream and dual-stream models. 
Single-stream models~\cite{Chen2020UNITERUI,visualbert,dalle,m6,vilt,shen2022how} concatenate the patch-wise or regional visual features and textual embeddings
and feed them into a single model. 
Dual-stream models \citep{vilbert,clip,align,lit,yao2022filip} use separate encoders for images and texts, allowing
 efficient inference for downstream multimodal alignment tasks like image-text retrieval, 
 by pre-computing image/text features offline. However, these models can not be directly used for multimodal generation tasks. 
In this paper, 
we  propose an efficient method to align the dual-stream VLP model CLIP's multimodal embedding space  with a powerful PLM BART to gain multimodal generation ability. 

There are also VLP models  that can perform multimodal generation tasks, by expensive pre-training with objective of image-conditioned auto-regressive language modeling~\cite{m6,simvlm,Hu2021ScalingUV,li2022blip}.
However, 
the pre-training of these models requires a large number of image-text pairs and numerous computation resources.
Other models like ~\cite{nocaps,visualbert,oscar,unifying,unimo} rely on an extra pre-trained object detector such as Faster-RCNN with labeled bounding-box data
to extract image regional features offline and are less scalable.

\subsection{Knowledge Distillation} 
Knowledge distillation (KD) in deep learning is first proposed by~\citet{hinton2015distilling}, which transfers knowledge embedded in the logits learned in a cumbersome teacher model to a smaller student model  without sacrificing too much performance.
Besides logits, other forms of knowledge like the intermediate representations and attentions~\cite{jiao2019tinybert,hou2020dynabert} have also been used in transferring the knowledge embedded in Transformer-based models. 
Recently, contrastive representation distillation \cite{tian2019contrastive} distills the knowledge from the teacher network to the student network by maximizing the mutual information between the two networks,
and is recently extended to transfer the knowledge from the pre-trained multimodal model CLIP
for zero-shot detection~\cite{gu2021zero} and multilingual setting~\cite{jain2021mural}.
In this paper, we apply the conventional KD as well as the contrastive KD to transfer the knowledge from the pre-trained CLIP to BART. Besides, we also propose to transfer the knowledge in CLIP image encoder to BART decoder through the cross-attention.

\begin{figure*}
    \centering
    \begin{subfigure}[b]{0.44\linewidth}
        \includegraphics[width=\linewidth]{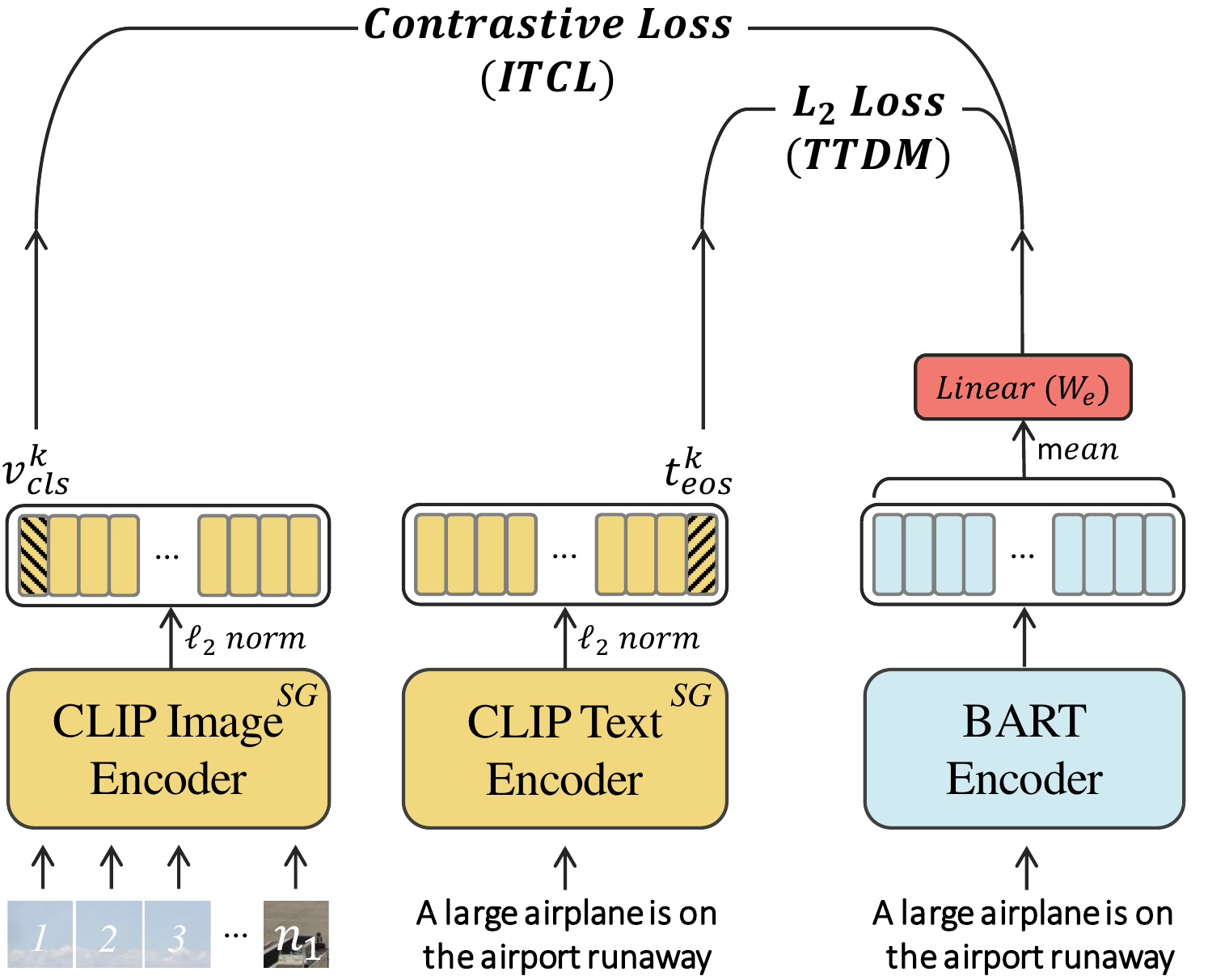}
        \caption{The \textit{TTDM} and \textit{ITCL} losses.}
        \label{fig:model_a}
    \end{subfigure}
    \hfill
    \begin{subfigure}[b]{0.45\linewidth}
        \centering
        \includegraphics[width=\linewidth,right]{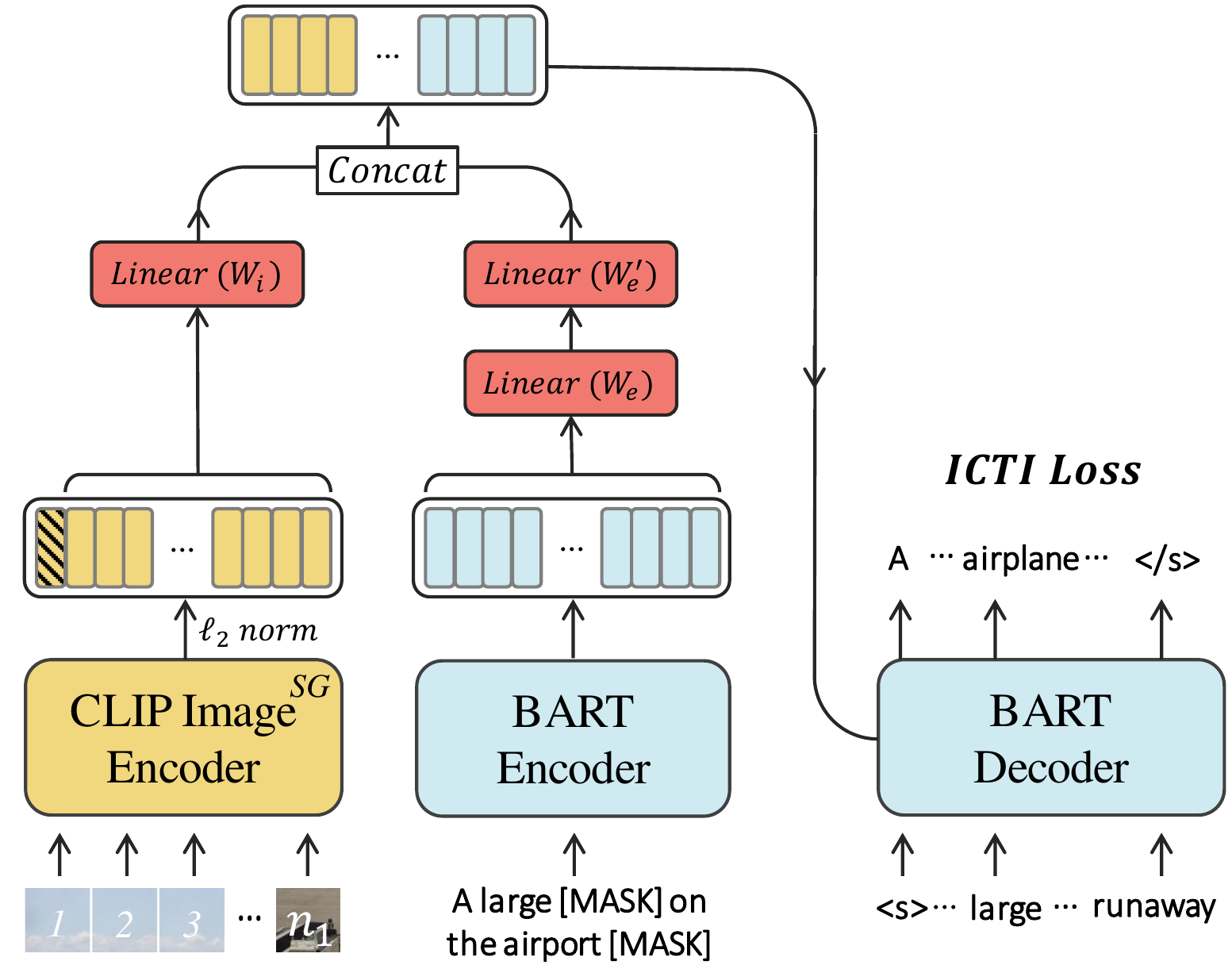}
        \caption{The \textit{ICTI} loss.}
        \label{fig:model_b}
    \end{subfigure}
    \caption{Architecture of the proposed VLKD method to distill multimodal knowledge from CLIP to BART.  (a) shows the \textit{TTDM} and \textit{ITCL} losses between the dual-stream CLIP encoders and BART encoder.  (b) illustrates the \textit{ICTI} loss for image-conditioned language modeling. \textit{SG} denotes the \textit{stop gradient} operation, indicating that no gradients will be back-propagated through that part of model parameters.}
    \label{fig:model}
\end{figure*}


\section{Proposed Method}
\label{sec:methodology}

We propose to distill multimodal knowledge from CLIP to BART for generative multimodal tasks, which takes the strengths from both sides (powerful multimodal representations of CLIP and text generation ability of BART). To this end, we propose three objectives (Section~\ref{sec:objectives}).
The overall architecture is illustrated in Figure~\ref{fig:model}.



\subsection{Model Architecture} \label{sec:model_architecture}

\paragraph{CLIP.} 
CLIP~\citep{clip} is a dual-stream VLP model pre-trained with a contrastive loss on 400 million image-text pairs. It consists of a text encoder which is a GPT~\citep{radford2019language} style Transformer model, and an image encoder which can be either a Vision Transformer (ViT)~\citep{vit} or Residual Convolutional Neural Network (ResNet)~\citep{resnet}. CLIP learns a joint multimodal embedding space with its text encoder and image encoder aligned.
Given an input image-text pair,
the ViT image encoder first reshapes the image 
into a sequence of 2D patches and then maps them into 1D embeddings with a prepended \texttt{[CLS]} token 
using a trainable linear projection. 
These embeddings are fed into the CLIP image encoder together with positional encodings.
The 
output embedding of the \texttt{[CLS]} token 
can represent the whole image.
When using ResNet-based image encoder, the \texttt{[CLS]} embedding is the average of output embeddings and then go through an attention pooling layer.
For the text sentence, 
it is bracketed with \texttt{[SOS]} and \texttt{[EOS]} tokens, and
the 
output embedding of the latter
is used as the sentence-level representation. In this paper, we explore four CLIP variants, including ViT-B/16, ViT-L/14, RN50$\times$16, and RN50$\times$64.



\paragraph{BART.} 
BART is a Transformer-based~\citep{transformer} sequence-to-sequence model that has a bi-directional encoder and a uni-directional (left-to-right) decoder, which can be seen as a generalization of the BERT~\citep{bert} and GPT~\citep{gpt}. It is pre-trained on 160GB text data in a self-supervised way by performing the text span infilling 
task with the input sentences corrupted and shuffled. 
Similar to the CLIP text encoder, 
BART also tokenizes and converts the input text into a sequence of embeddings,
which are then fed into the BART encoder. 
BART excels at 
both
NLG 
(e.g., abstractive summarization)
and
NLU tasks. 

\subsection{Training Objectives} \label{sec:objectives}
To distill multimodal knowledge from CLIP to BART, we propose three objective functions: 
1) Text-Text Distance Minimization (\textit{TTDM}); 
2) Image-Text Contrastive Learning (\textit{ITCL}); and 
3) Image-Conditioned Text Infilling (\textit{ICTI}).
During training, the model parameters of CLIP are frozen constantly, i.e. no gradients will be back-propagated through them (marked as \textit{SG} in Figure~\ref{fig:model}), to ensure its two encoders are still aligned and the multimodal knowledge is not forgotten.

For each training batch with $B$ image-text pairs,
denote the $k$-th image-text pair as $\x^k = \{\x^k_I, \x^k_T\}$, and the output of multimodal encoders of CLIP and BART encoder as 
\begin{eqnarray*}
    \text{CLIP}_{I}(\x^k_I) \!\!\!&\rightarrow &\!\!\! \V^k = [ \v^k_{cls}, \v^k_1, \dots, \v^k_{n_1} ], \\
    \text{CLIP}_{T}(\x^k_T) \!\!\!&\rightarrow &\!\!\! \T^k = [ \t^k_{sos}, \t^k_1, \dots, \t^k_{n_2}, \t^k_{eos} ], \\
    \text{BART}_{enc}(\x^k_T) \!\!\!& \rightarrow &\!\!\! \E^k = [ \e^k_{bos}, \e^k_1, \dots, \e^k_{n_3}, \e^k_{eos} ].
\end{eqnarray*}
Here, $n_1$ is the number of image patches, $n_2$ and $n_3$ denote the  sequence lengths of the text encoder of CLIP and BART, respectively.
$\v^k_{*}, \t^k_{*} \in \mathbb{R}^{d_1}$ represents the $\ell_2$-normalized output embedding from the CLIP image and text encoder
at a certain position.
$\e^k_{*}$ is the unnormalized raw output embedding from the BART encoder.
In the following, we elaborate on the three distillation objectives.

\subsubsection{Text-Text Distance Minimization}
\label{sec:ttdm}
To align the CLIP text encoder and BART encoder, i.e. making their output representations
close given the same input text, 
we propose to minimize the $\ell_2$ distance between their sequence-level output representations. 
Specifically, for the $k$-th input text, it can be formulated as 
\begin{gather*}
    \bar \e^k_{\text{norm}} = \W_e \bar \e^k /\| \W_e \bar \e^k \|_2, \\
    \mathcal{L}_{TTDM} = \frac{1}{B}\sum^B_{k=1}\| \t^k_{eos} - \bar \e^k_{\text{norm}}\|^2,
\end{gather*}
where $\bar \e^k \in \mathbb{R}^{d_2}$ is the average of all output embeddings from the BART encoder, and $\W_e \in \mathbb{R}^{d_1 \times d_2}$ is a weight matrix 
to linearly project the output of BART encoder to CLIP's multimodal space.

\subsubsection{Image-Text Contrastive Learning} 
\label{sec:itcl}
Contrastive training has been shown to be very effective in cross-modal representation learning~\citep{tian2020contrastive,sigurdsson2020visual,zhang2020contrastive,clip}. 
To further adapt the BART encoder to CLIP's multimodal space, 
we optimize a symmetric InfoNCE loss between the output representations of the BART encoder and CLIP image encoder. 
The image-to-text contrastive loss $\mathcal{L}_{i2t}$ is formulated as
\begin{gather*}
    \mathcal{L}_{i2t} = -\frac{1}{B} \sum^B_{k=1} \log \frac{\exp(\v^{k\top}_{cls} \bar \e^k_{\text{norm}} / \tau)}{\sum_j \exp(\v^{k\top}_{cls}  \bar \e^j_{\text{norm}} / \tau)},
\end{gather*}
where
$\tau$ is a learnable temperature parameter. Different from \citet{clip}, we find that not clamping the $\tau$ shows a slight improvement. 
Similarly, the text-to-image contrastive loss
$\mathcal{L}_{t2i}$ is
\begin{gather*}
    \mathcal{L}_{t2i} = -\frac{1}{B} \sum^B_{k=1} \log \frac{\exp(\v^{k\top}_{cls} \bar \e^k_{\text{norm}} / \tau)}{\sum_j \exp(\v^{j\top}_{cls} \bar \e^k_{\text{norm}} / \tau)}.
\end{gather*}
Then, the 
\textit{ITCL} 
loss can be calculated as 
\begin{gather*}
    \mathcal{L}_{ITCL} = \frac{1}{2}(\mathcal{L}_{i2t} + \mathcal{L}_{t2i}).
\end{gather*}

Note that when computing the \textit{ITCL} and \textit{TTDM} losses, we do not introduce any new linear projections to the CLIP output features to avoid destroying the pre-trained alignment between its image and text encoders.
Instead, we add one linear layer (parameterized by $\W_e$) to project the BART encoder to CLIP's representation space and match their feature dimension.


\subsubsection{Image-Conditioned Text Infilling} 
\label{sec:icti}
With only \textit{TTDM} and \textit{ITCL}, the BART decoder is not updated at all. 
To consort BART encoder and decoder, we propose to perform the text span infilling task conditioned on the corresponding image features. 
As depicted in Figure~\ref{fig:model_b}, for the $k$-th image-text pair, following \citet{bart},
we corrupt the input text by masking 15\% of whole-word tokens with span lengths drawn from a Poisson Distribution with $\lambda=3$. 

Considering that $\V^k$ and $\W_e \E^k$ are already aligned in the CLIP's multimodal space through \textit{TTDM} and \textit{ITCL}, and having a different feature dimension with the BART decoder, we further project them to the BART decoder dimension with $\W_i$ and $\W'_e$.
Then, we concatenate them together as $\C^k$ before feeding into the BART decoder as shown in Eq.\eqref{eq:3}. 
As mentioned in Section~\ref{sec:model_architecture}, we explore two variants of CLIP. 
With a slight abuse of notation,
for the ResNet-based CLIP, $\V^k$ is composed of all embeddings after the final attention pooling layer $\{\v^k_{i}\}_{i=1}^{n_1}$, 
while for the ViT-based CLIP, $\V^k$ consists of the embedding of the \texttt{[CLS]} token $\v^k_{cls}$ only. 

Note that the weight matrix $\W'_e$ is initialized to be the pseudo-inverse of $\W_e$, such that text representations after the two projections $\W'_e \W_e \E^k$ are the closest to the original pre-trained BART encoder space at initialization\footnote{The pseudo inverse matrix $\W'_e$ satisfies  $\W'_e =  \arg \min_{\X}\|\W_e \X - \I\|_F^2$, where $\I$ is the identity matrix and $\|\cdot\|_F$ denotes the Frobenius Norm.}. 
The BART decoder then interacts with 
$\C^k$ through standard Transformer cross-attention layers. We optimize a language modeling loss $\mathcal{L}_{ICTI}$ by minimizing the negative log-likelihood in Eq.\eqref{eq:icti}, in which $\w_j$ denotes the token to be predicted at each decoding step.
\begin{gather}
    \C^k = \text{concat} (\W_i \V^k, \W_e'\W_e \E^k), \label{eq:3} \\
    \mathcal{L}_{ICTI} = - \frac{1}{B} \sum^B_{k=1} \sum_j \log P(\w_j^k | \w_{<j}^k, \C^k). \label{eq:icti}
\end{gather}

The \textit{ICTI} loss is crutial for
for our methodology to work, as it not only coordinates the BART encoder and decoder, but also enables the BART decoder to understand the multimodal information by recovering texts with visual clues. 

Finally, we simultaneously optimize the 
summation of three losses $\mathcal{L}$ as 
\begin{gather*}
    \mathcal{L} = \gamma \mathcal{L}_{TTDM} + \mathcal{L}_{ITCL} + \mathcal{L}_{ICTI},
\end{gather*}
where $\gamma$ is set to $10^3$ by default, as $\mathcal{L}_{ITCL},\mathcal{L}_{ICTI}$ are about 
three magnitudes
larger than $\mathcal{L}_{TTDM}$. 


\subsection{Datasets for VLKD} \label{sec:dataset}
Our model is trained on the Conceptual Captions (CC3M)~\citep{cc3m} dataset, which contains 3 million image-text pairs crawled from the Internet. 
For larger model variants (ViT-L/14 and RN50x64), we further include the Visual Genome Caption data which contains $\sim$700K image-text pairs. No images for pre-training appear in the downstream datasets.
Compared to previous VLP work~\citep{clip,align,simvlm},
VLKD is much cheaper by leveraging several magnitudes less data. Furthermore, we experiment with even smaller data (1M, 100K) by uniformly sampling a subset of CC3M to test the limit of dataset size of VLKD, 
with results
discussed in Section~\ref{sec:ablation_study}.


\section{Experiments} \label{sec:experiments}
To demonstrate the effectiveness of VLKD, we evaluate it on generative multimodal tasks for both zero-shot and finetuning.
Specifically, we test the image captioning task, and also the VQA task under the open-ended scenario. Furthermore, we also run the model on NLU and NLG tasks to investigate the influence of VLKD on the text processing ability of the original pre-trained BART.



\subsection{Finetuning Datasets}

\paragraph{Image Captioning.} Image captioning requires the model to generate a relevant description given an image. We use the COCO image caption dataset~\citep{Lin2014MicrosoftCC} with the Karpathy split~\citep{Karpathy2017DeepVA}. Additionally, we use the NoCaps~\citep{nocaps} dataset to test the model performance when there are out-of-domain 
objects.

\paragraph{Open-Ended VQA.} 
Unlike previous works~\citep{Anderson2018BottomUpAT,Chen2020UNITERUI,oscar,Yu2021ERNIEViLKE,vinvl,Kim2021ViLTVT} that treat the VQA task as a discriminative problem, we let the model generate answers freely, which is more aligned with the real-world scenario of this task. We use the standard VQAv2~\citep{vqav2}, and also OK-VQA~\citep{okvqa} which requires knowledge to answer questions correctly.

\paragraph{NLU and NLG.} 
For NLU, we test our model on the GLUE benchmark~\citep{wang2018glue}, which consists of nine text classification tasks. We exclude the WNLI task as it is problematic\footnote{\url{https://gluebenchmark.com/faq}
}. For NLG, we test the abstractive summarization task on  XSUM~\citep{xsum-emnlp} dataset, which requires the model to comprehend long texts and generate short summaries with key information.


\subsection{Implementation Details}
We use BART-large as the pre-trained backbone NLP model, which has 12 layers in both encoder and decoder with a hidden size of 1024 and 16 heads in each multi-head attention (MHA) layer. 
In total, it contains 406M parameters. 
For the pre-trained CLIP~\cite{clip} model, we report four variants with different visual backbones, including ViT-B/16, ViT-L/14, RN50$\times$16, and RN50$\times$64. 

We use 64 Nvidia V100 GPUs for VLKD and 8 for the finetuning of downstream tasks. In total, we pre-train the model for 10 epochs, which takes about 5 hours. We use a batch size of 4608 for ViT-B/16 and ViT-L/14, 4096 for RN50x16 and 3840 for RN50x64. All of the models are optimized by the AdamW~\citep{adamw} optimizer. The learning rate is warmed up to $2.4e^{-4}$ within the first 2\% steps and then linearly decay to 0.
More information of VLKD pre-training and the finetuning of each downstream task can be found in Appendix \ref{sec:appendix_hyperparams}.




\begin{figure*}[htbp]
    \centering
    \begin{subfigure}[b]{0.49\linewidth}
        \centering
        \includegraphics[width=\linewidth]{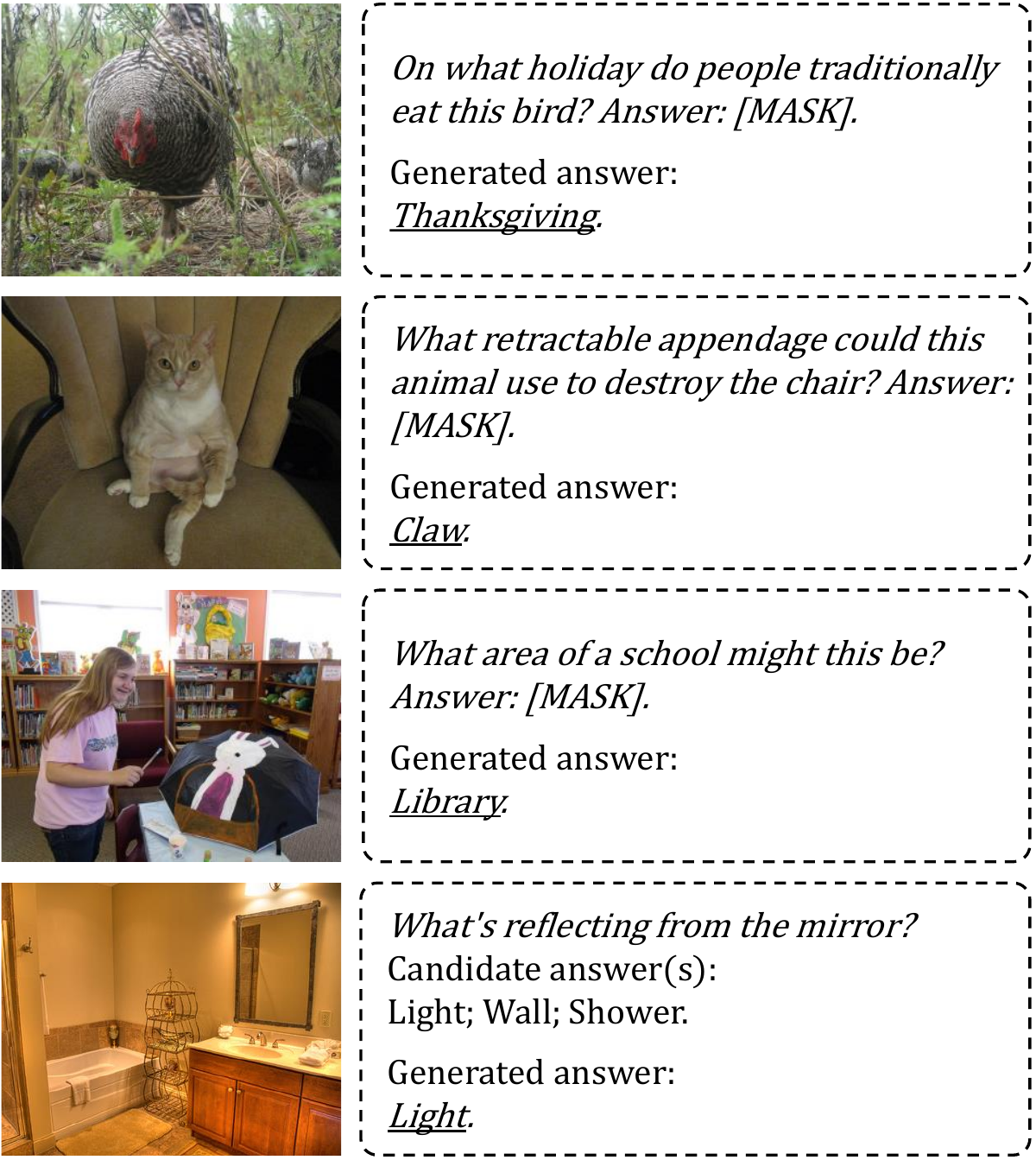}
        \caption{Zero-shot VQA.}
        \label{fig:case_study_a}
    \end{subfigure}
    \hfill
    \begin{subfigure}[b]{0.49\linewidth}
        \centering
        \includegraphics[width=\linewidth,right]{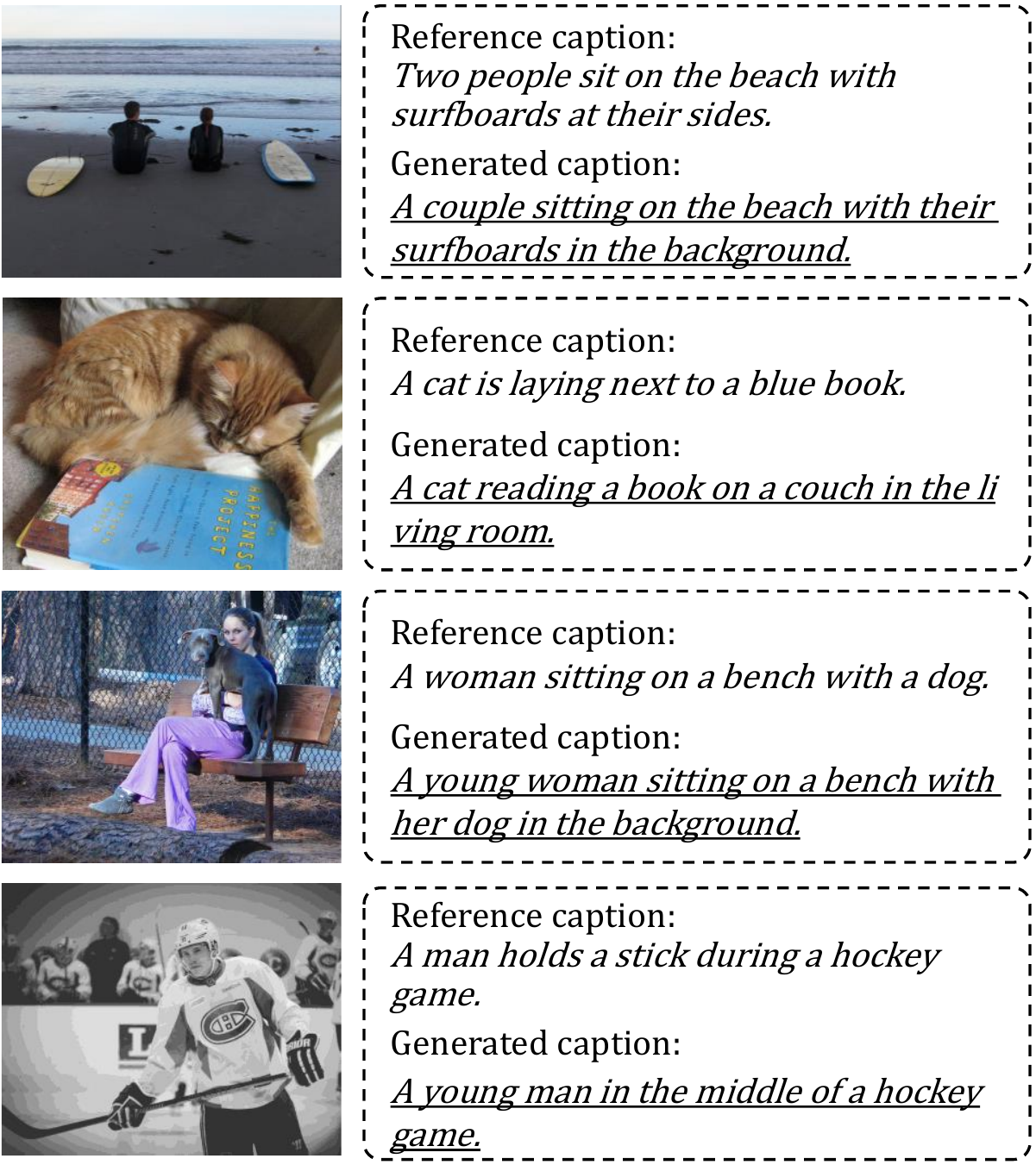}
        \caption{Zero-shot image captioning.}
        \label{fig:case_study_b}
    \end{subfigure}
    \caption{Examples of 
    (a) zero-shot VQA and (b) image captioning.
    Our model shows the ability 
    to recognize
    visual objects and generate appropriate sentences based on their properties and relationship. Furthermore, the model can bind visual objects to text conceptual knowledge that is learned in the PLMs when generating answers given questions.}
    \label{fig:case_study}
\end{figure*}

\subsection{Multimodal Zero-Shot Evaluation}
Benefit from the knowledge distillation, especially the \textit{ICTI} loss, our model can perform various downstream multimodal tasks in a zero-shot manner. 

\subsubsection{Zero-Shot Image Captioning} \label{sec:zs_image_caption}
During knowledge distillation, the \textit{ICTI} loss can be seen as a simple version of the image captioning task, which asks the model to fill in the corrupted locations of image descriptions. 
If the masking ratio increases to 100\%, it reduces to the image captioning task. 
Therefore, it is intuitive to test the zero-shot performance of our model.

\begin{table*}[t]
\centering
\resizebox{\textwidth}{!}{
\begin{tabular}{l|c|cc|cccc|cccccccc}
\toprule
\multirow{3}{*}{Methods} & \multirow{3}{*}{\begin{tabular}[c]{@{}c@{}}\#Pretrain\\Image-text\\Pairs\end{tabular}} & \multirow{3}{*}{OD} & \multirow{3}{*}{OT} & \multicolumn{4}{c|}{\multirow{2}{*}{\begin{tabular}[c]{@{}c@{}}COCO Caption\\ Karpathy Test\end{tabular}}} & \multicolumn{8}{c}{NoCaps Validation} \\
 &  &  && \multicolumn{4}{c|}{} & \multicolumn{2}{c}{In} & \multicolumn{2}{c}{Near} & \multicolumn{2}{c}{Out} & \multicolumn{2}{c}{Overall} \\
 &  &&  & \multicolumn{1}{c}{B@4} & \multicolumn{1}{c}{C} & \multicolumn{1}{c}{M} & \multicolumn{1}{c|}{S} & C & S & C & S & C & S & C & S \\ \midrule \midrule
$\text{BUTD}^\dagger$  & 1.5M & \ding{51} & \ding{51} & \multicolumn{1}{c}{36.3} & \multicolumn{1}{c}{120.1} & \multicolumn{1}{c}{27.7} & \multicolumn{1}{c|}{21.4} & 80.0 & 12.0 & 73.6 & 11.3 & 66.4 & 9.7 & 73.1 & 11.1 \\
${\text{OSCAR}_{\text{Large}}^\dagger}$ & 6.5M & \ding{51} & \ding{51} & \multicolumn{1}{c}{41.7} & \multicolumn{1}{c}{140.0} & \multicolumn{1}{c}{30.6} & \multicolumn{1}{c|}{24.5} & 85.4 & 11.9 & 84.0 & 11.7 & 80.3 & 10.0 & 83.4 & 11.4 \\ 
$\text{VinVL}_{\text{Large}}$ & 6.5M & \ding{51} & \ding{51} & \multicolumn{1}{c}{41.0} & \multicolumn{1}{c}{140.9} & \multicolumn{1}{c}{31.1} & \multicolumn{1}{c|}{25.2} & 103.7 & 13.7 & 95.6 & 13.4 & 83.8 & 11.9 & 94.3 & 13.1 \\
VL-T5  & 9.2M & \ding{51} & \ding{55} & 34.6 & 116.1 & 28.8 & 21.9 & - & - & - & - & - & - & - & - \\
VL-BART &  9.2M & \ding{51} & \ding{55} & 34.2 & 114.1 & 28.4 & 21.3 & - & - & - & - & - & - & - & - \\ 
\textcolor{tablegray}{$\text{LEMON}_{\text{Huge}}$} & \textcolor{tablegray}{203M} & \textcolor{tablegray}{\ding{51}} & \textcolor{tablegray}{\ding{51}} & \textcolor{tablegray}{42.6} & \textcolor{tablegray}{145.5} & \textcolor{tablegray}{31.4} & \textcolor{tablegray}{25.5} & \textcolor{tablegray}{118.0} & \textcolor{tablegray}{15.4} & \textcolor{tablegray}{116.3} & \textcolor{tablegray}{15.1} & \textcolor{tablegray}{120.2} & \textcolor{tablegray}{14.5} & \textcolor{tablegray}{117.3} & \textcolor{tablegray}{15.0} \\
\textcolor{tablegray}{$\text{SIMVLM}_{\text{Huge}}$} & \textcolor{tablegray}{1.8B} & \textcolor{tablegray}{\ding{55}} & \textcolor{tablegray}{\ding{55}} & \textcolor{tablegray}{40.6} & \textcolor{tablegray}{143.3} & \textcolor{tablegray}{33.7} & \textcolor{tablegray}{25.4} & \textcolor{tablegray}{113.7} & \textcolor{tablegray}{-} & \textcolor{tablegray}{110.9} & \textcolor{tablegray}{-} & \textcolor{tablegray}{115.2} & \textcolor{tablegray}{-} & \textcolor{tablegray}{112.2} & \textcolor{tablegray}{-} \\ \midrule
VLKD \small{\textit{({\text{Zero-shot}})}} & &  &  &  &  &  &  &  &  &  &  &  &  &  &  \\
\small{ViT-B/16}  & 3M & \ding{55} & \ding{55} & 16.7 & 58.3 & 19.7 & 13.4 & - & - & - & - & - & - & - & - \\
\small{RN50$\times$16}  & 3M & \ding{55} & \ding{55} & 18.2 & 61.1 & 20.8 & 14.5 & 52.6 & 9.7 & 52.9 & 9.6 & 58.6 & 9.3 & 54.0 & 9.6 \\
\small{RN50$\times$64}  & 3.7M & \ding{55} & \ding{55} & 25.8 & 85.1 & 23.1 & 16.9 & 64.8 & 13.6 & 62.3 & 13.6 & 66.9 & 9.9 & 63.6 & 12.8 \\ \midrule
VLKD \small{\textit{({\text{Finetuned}})}} & &  &  &  &  &  &  &  &  &  &  &  &  &  &  \\
\small{ViT-B/16}  & 3M & \ding{55} & \ding{55} & 37.2 & 128.0 & 28.8 & 22.4 & - & - & - & - & - & - & - & - \\
\small{RN50$\times$16}  & 3M & \ding{55} & \ding{55} & 38.9 & 131.1 & 29.6 & 23.9 & 92.3 & 12.6 & 82.0 & 11.8 & 70.3 & 10.4 & 81.1 & 11.7 \\
\small{RN50$\times$64}  & 3.7M & \ding{55} & \ding{55} & 40.3 & 135.7 & 30.5 & 24.3 & 105.1 & 14.5 & 99.7 & 13.8 & 90.2 & 12.1 & 97.6 & 13.6 \\ \bottomrule
\end{tabular}
}
\caption{Results on the COCO caption (Karpathy test set) and NoCaps (validation set). B@4, C, M, and S denote BLEU-4, CIDEr, METEOR, and SPICE, respectively. OD and OT indicate whether object detectors and object tags are used or not. Numbers of previous models are taken from~\cite{Anderson2018BottomUpAT,oscar,vinvl,unifying,Hu2021ScalingUV,simvlm}. Models marked by $\dagger$ additionally use the constrained beam search (CBS)~\citep{cbs} for the NoCaps dataset. Note that LEMON and SIMVLM use significantly more pre-training data and have more trainable model parameters than the others.}
\label{tab:caption}
\end{table*}

\begin{table}[h!]
\centering
\resizebox{\linewidth}{!}{%
\begin{tabular}{lccc}
\toprule
\multicolumn{1}{l|}{Methods} & \multicolumn{1}{c|}{\#Params} & \begin{tabular}[c]{@{}c@{}}VQAv2\\ val / test-dev\end{tabular} & \begin{tabular}[c]{@{}c@{}}OK-VQA\\ test\end{tabular} \\ \midrule \midrule
\multicolumn{4}{c}{\textit{Generative (Open-ended)}} \\ \midrule 
\multicolumn{1}{l|}{Frozen \small{\textit{(Zero-shot)}}} & \multicolumn{1}{c|}{\multirow{2}{*}{7B}} & 29.5 / - & 5.9 \\
\multicolumn{1}{l|}{Frozen \small{\textit{(Finetuned)}}} & \multicolumn{1}{c|}{} & 48.4 / - & 19.6 \\ \midrule
\multicolumn{1}{l|}{VLKD \small{\textit{({\text{Zero-shot}})}}} & \multicolumn{1}{c|}{\multirow{8}{*}{< 1B}} &  &  \\
\multicolumn{1}{l|}{\small{RN50$\times$16}} & \multicolumn{1}{c|}{} & 37.4 / 38.2 & 9.9 \\
\multicolumn{1}{l|}{\small{ViT-B/16}} & \multicolumn{1}{c|}{} & 38.6 / 39.7 & 10.5 \\
\multicolumn{1}{l|}{\small{ViT-L/14}} & \multicolumn{1}{c|}{} & 42.6 / 44.5 & 13.3 \\ \cmidrule{1-1} \cmidrule{3-4} 
\multicolumn{1}{l|}{VLKD \small{\textit{({\text{Finetuned}})}}} & \multicolumn{1}{c|}{} &  &  \\
\multicolumn{1}{l|}{\small{RN50$\times$16}} & \multicolumn{1}{c|}{} & 67.4 / 68.8 & 36.2 \\
\multicolumn{1}{l|}{\small{ViT-B/16}} & \multicolumn{1}{c|}{} & 69.3 / 69.8 & 36.3 \\
\multicolumn{1}{l|}{\small{ViT-L/14}} & \multicolumn{1}{c|}{} & 73.9 / 74.5 & 39.0 \\ \midrule
\multicolumn{4}{c}{\textit{Discriminative}} \\ \midrule
\multicolumn{1}{l|}{$\text{UNITER}_{\text{Large}}$} & \multicolumn{1}{c|}{-} & - / 73.8 & - \\
\multicolumn{1}{l|}{$\text{OSCAR}_{\text{Large}}$} & \multicolumn{1}{c|}{-} & - / 73.6 & - \\
\multicolumn{1}{l|}{$\text{VinVL}_{\text{Large}}$} & \multicolumn{1}{c|}{-} & - / 76.5 & - \\ 
\multicolumn{1}{l|}{\textcolor{tablegray}{$\text{SIMVLM}_{\text{Base}}$}} & \multicolumn{1}{c|}{\textcolor{tablegray}{-}} & \textcolor{tablegray}{- / 77.9} & \textcolor{tablegray}{-} \\ \bottomrule
\end{tabular}%
}
\caption{Accuracies(\%) on the VQAv2 and OK-VQA datasets. We categorize models into two parts: answer questions in a generative or discriminative way.}
\label{tab:vqa}
\end{table}

\begin{table}[htbp]
\centering
\scalebox{0.86}{
\setlength{\tabcolsep}{2.8mm}{
\begin{tabular}{l|cc}
\toprule
Model   & In-domain & Out-of-domain \\ \midrule \midrule
UNITER  & \textbf{74.4}      & 10.0       \\ \midrule
VL-T5   & 71.4      & 13.1       \\
VL-BART & 72.1      & 13.2       \\
VLKD (ViT-L/14)    & 74.9      & \textbf{23.4}       \\ \bottomrule
\end{tabular}%
}
}
\caption{Accuracies(\%) on VQAv2 Karpathy test-split.}
\label{tab:vqa_karpathy_test}
\end{table}

Following \citet{clip} and \citet{simvlm}, we compose the input with a text prompt and also
$m$ mask tokens, i.e., ``A picture of \texttt{[MASK]$\times m$}.'', for the model to generate the caption for the image.
The zero-shot results are included in Table~\ref{tab:caption}. 
Our zero-shot model achieves comparable overall performance to the finetuned UpDown~\citep{nocaps} model on NoCaps dataset.
As shown in Figure~\ref{fig:case_study_b}, the zero-shot generated captions are plausible with correct objects, relationships, and actions. 
However, sometimes details like colors could be omitted. 

In our experiments, we use $m=6$ for COCO and $m=8$ for NoCaps. Although it could potentially limit the length of generation, we find that it has negligible influence to the performance, as for each \texttt{[MASK]} token, the model is learned to fill one to three tokens depending on the context. Furthermore, this could be used to control the length of generated texts for different senarios. See Section~\ref{sec:ablation_study} for a more detailed discussion about the effects of number of the masks.

\subsubsection{Zero-Shot VQA}
Zero-shot VQA is much more challenging than image captioning, as it requires reasoning over both the image and question, which is very different from the \textit{ICTI} loss during the knowledge distillation. 
As illustrated in Figure~\ref{fig:intro}, we construct the input by appending a text prompt ``\text{Answer:} \texttt{[MASK]$\times n$}.'' to the question
Given the context (image+question+prompt), the model is required to 
predict the answer by \textit{recovering} the textual token in the \texttt{[MASK]} positions. In our experiments, 
we use $n=2$ for the VQAv2, which is found performing best among $n \in \{1, 2, 3\}$.

In Table~\ref{tab:vqa}, compared to the strong baseline Frozen~\citep{Tsimpoukelli2021MultimodalFL}, our model improves the zero-shot accuracy by 13.1\% on the VQAv2 validation set and 7.4\% on the OK-VQA test set with 7$\times$ fewer  parameters, indicating the efficiency and effectiveness of VLKD. Our model achieves 44.5\% zero-shot accuracy on the VQAv2 test-dev set, which to the best of our knowledge is the new state-of-the-art.
Furthermore, as shown in Figure~\ref{fig:case_study_a}, our model can bind visual objects to conceptual knowledge stored in the PLM to answer questions. For example, it connects the visual object \textit{Turkey} with the traditional food people usually eat at the \textit{Thanksgiving} festival.

\subsection{Multimodal Finetuning Evaluation}
When finetuning VLKD on downstream multimodal tasks, we keep the same input format as zero-shot to obtain outputs in a generative way. The CLIP model parameters are still frozen during finetuning.

\begin{table*}[t]
\centering
\resizebox{\textwidth}{!}{%
\begin{tabular}{lcccccccc}
\hline \toprule
      Model & CoLA & SST-2 & RTE  & MRPC      & QQP       & MNLI & QNLI & Avg. \\ \midrule \midrule
\multicolumn{1}{l|}{$\text{BERT}_{\text{LARGE}}^\diamond$ \citep{bert}}       & 60.6 & 93.2  & 70.4 & 82.9/88.0 & 91.3/87.9 & 86.4 & 92.3 & 82.6 \\ 
\multicolumn{1}{l|}{$\text{BART}_{\text{LARGE}}^\diamond$ \citep{bart}}       & \textbf{62.8} & \textbf{96.6}  & \textbf{87.0} & 86.7/90.4 & \textbf{92.5/89.3} & \textbf{90.0} & \textbf{94.9} & \textbf{87.2} \\ 
\midrule
\multicolumn{1}{l|}{$\text{VisualBERT}^\dagger$ \citep{visualbert}} & 38.6 & 89.4  & 56.6 & 71.9/82.1 & 89.4/86.0 & 81.6 & 87.0 & 74.0 \\
\multicolumn{1}{l|}{$\text{UNITER}^\dagger$ \citep{Chen2020UNITERUI}}     & 37.4 & 89.7  & 55.6 & 69.3/80.3 & 89.2/85.7 & 80.9 & 86.0 & 73.1 \\
\multicolumn{1}{l|}{$\text{VL-BERT}^\dagger$ \citep{vlbert}}    & 38.7 & 89.8  & 55.7 & 70.6/81.8 & 89.0/85.4 & 81.2 & 86.3 & 73.6 \\
\multicolumn{1}{l|}{$\text{VilBERT}^\dagger$ \citep{vilbert}}    & 36.1 & 90.4  & 53.7 & 69.0/79.4 & 88.6/85.0 & 79.9 & 83.8 & 72.1 \\
\multicolumn{1}{l|}{$\text{LXMERT}^\dagger$ \citep{lxmert}}     & 39.0 & 90.2  & 57.2 & 69.8/80.4 & 75.3/75.3 & 80.4 & 84.2 & 71.6 \\
\multicolumn{1}{l|}{$\text{SIMVLM}^\ddagger$ \citep{simvlm}}     & 46.7 & 90.9  & 63.9 & 75.2/84.4 & 90.4/87.2 & 83.4 & 88.6 & 77.4 \\ \midrule
\multicolumn{1}{l|}{VLKD (RN50$\times$16)}       & 59.1 & 95.5  &   81.2  & \textbf{87.5/91.1} &     92.1/89.2      &   89.6   & 94.3 &   85.7   \\ \bottomrule
\end{tabular}%
}
\caption{
Results on the GLUE development set
(single task single models). We report the Matthews correlation for CoLA, accuracy/F1 for MRPC and QQP, and accuracy for the rest of the tasks. The performance of models that are marked by \(\diamond\) are taken from~\citep{bart}, \(\dagger\) are from \citep{Iki2021EffectOV}, and \(\ddagger\) are from \citep{simvlm}. Compared to other VLP models, our VLKD model has a great advantage in text-only NLP tasks.}
\label{tab:glue}
\end{table*}

\subsubsection{Finetuning Image Captioning} 
In Table~\ref{tab:caption}, we demonstrate that our model can achieve decent performance when finetuned on the COCO dataset. The SCST CIDEr optimization method~\cite{Rennie2017SelfCriticalST} is used to further improve the performance.
Our model outperforms VL-T5/BART~\citep{unifying} without using an extra object detector, which is fairly time-consuming as explained by \citet{Kim2021ViLTVT}. 
Compared to state-of-the-art models, however, there is still a small performance gap, which we conjecture is mainly due to their usage of object detector/tags and much more pre-training image-text pairs. 
We also evaluate our VLKD models with ResNet visual backbones on the NoCaps dataset (Table~\ref{tab:caption}). 
For zero-shot image caption, the CIDEr score on the out-of-domain set is even higher than the in- and near-domain sets, which shows the generalization of our knowledge distillation method to common visual objects. After finetuned on the COCO training set, the performance on NoCaps of our model with the RN50$\times$64 backbone is comparable to the state-of-the-art models.

\subsubsection{Finetuning VQA}

From Table~\ref{tab:vqa}, 
the best performance of VQAv2 is achieved by VLP models that tackle this task in a discriminative way with a set of pre-defined answers. However, this approach does not generalize to real-world scenarios and cannot be directly applied to more diverse datasets (e.g., OK-VQA). 
Differently, Frozen~\citep{Tsimpoukelli2021MultimodalFL} and our proposed VLKD formulate VQA as a generative problem to generate
answers conditioned on the questions and images in an open-ended manner, which also enables zero-shot VQA. Specifically, for each question-answer pair in the VQAv2 dataset, we optimize the model to generate the answer with the cross-entropy loss and a label-smoothing of 0.1. The loss is weighted by the weight of each answer candidate. In addition, we augment the training data with VG-QA~\cite{Krishna2016VisualGC}.


Furthermore, following \cite{unifying}, we test the performance on out-of-domain questions with rare answers using the Karpathy test-split. As shown in Table~\ref{tab:vqa_karpathy_test}, our method shows a salient advantage on out-of-domain questions 
due to the
benefit from VLKD and its generative nature without defining the answer list.


\begin{table}[t]
\centering
\scalebox{0.9}{
\begin{tabular}{l|ccc}
\toprule
Model  & ROUGE-1   & ROUGE-2   & ROUGE-L   \\ \midrule\midrule
$\text{BART}_{\text{Large}}$ & 45.14     & 22.27     & 37.25     \\
VLKD      & 44.86     & 22.06     & 36.95     \\ \bottomrule
\end{tabular}%
}
\caption{Results of abstractive summarization on XSUM. We use the best performing checkpoint of the RN50$\times$16 variant.}
\label{tab:summarization}
\end{table}

\subsection{Evaluation of NLU and NLG}

Table~\ref{tab:glue} shows results on the GLUE benchmark.
Although prior VLP models are either initialized from the pre-trained BERT model, or trained by a text-only language modeling loss together with the vision-language (VL) losses, they generally suffer from the weakened performance of NLU.
For example, SIMVLM performs significantly worse than BART, though trained with four times more textual data.
We speculate that
the weakened NLU ability of these models is caused by the
catastrophic forgetting of the language knowledge in the pre-trained BERT weights during the multimodal pre-training.
Moreover, 
simultaneous optimization of multimodal and text-only objectives potentially shifts the latter to be an auxiliary loss, making the NLP ability not as effective. 

On the other hand, the resulting model of VLKD performs only slightly worse than the original BART and significantly outperforms BERT, as the original knowledge embedded in BART is well maintained.

Additionally, as presented in Table~\ref{tab:summarization}, we also run VLKD on the abstractive summarization task to evaluate its NLG performance, since BART-based methods excel on the summarization~\cite{bart,gsum,yu-etal-2021-vision}. The gap between VLKD and its backbone BART is negligible. Overall, we empirically demonstrate that VLKD enables the backbone PLM to perform multimodal tasks without hurting its original NLP ability.


\section{Ablation Study} \label{sec:ablation_study}

\paragraph{Knowledge Distillation Objectives.}
Table~\ref{tab:kd_ablation} shows the ablation on the knowledge distillation objectives, except the \textit{ICTI} loss which
 is necessary for our method to work.
Without \textit{TTDM} or \textit{ITCL}, we observe a clear degradation of zero-shot performance on both VQAv2 and COCO image caption datasets. 
It is worth noting that \textit{ITCL} contributes more to the image captioning task, which requires a deeper perception of visual features to generate captions. 
Oppositely, \textit{TTDM} helps more for the VQA task, which involves reasoning over the question and image features. 
Removing both of them incurs a large performance drop, which demonstrates the importance of aligning the embedding space between CLIP and BART. 
\begin{table}[!h]
\centering
\resizebox{\linewidth}{!}{%
\begin{tabular}{l|cc}
\toprule
Model    & \begin{tabular}[c]{@{}c@{}}VQAv2 (val)\end{tabular} & \begin{tabular}[c]{@{}c@{}}COCO Caption (test)\end{tabular} \\ \midrule\midrule
$\text{VLKD}_{\text{ZERO-SHOT}}^{\text{ViT-B/16}}$ & 38.6      & 58.3    \\
$\;\;\text{w/o \textit{TTDM}}$ & 35.5    & 55.7   \\
$\;\;\text{w/o \textit{ITCL}}$ & 36.3    & 54.1    \\
$\;\;\text{w/o \textit{Both}}$ & 30.1    & 48.6    \\ \bottomrule
\end{tabular}%
}
\caption{Ablation study on three distillation objectives.}
\label{tab:kd_ablation}
\end{table}

\paragraph{Number of Masks.}
Furthermore, we also test the influence of the number of masks for zero-shot image captioning in Table~\ref{tab:ablation_masks}. As discussed in Section~\ref{sec:zs_image_caption}, it has a trivial influence as the model learns to fill a variable length of tokens for each masked position. We achieve the best performance on the COCO caption dataset when $m=6$ and NoCaps when $m=8$.
\begin{table}[h]
\centering
\scalebox{0.9}{
\setlength{\tabcolsep}{4.3mm}{
\begin{tabular}{l|cccc}
\toprule
\#masks & 5 & 6 & 7 & 8 \\ \midrule \midrule
CIDEr & 59.7 & 61.1 & 60.6 & 59.6 \\ \bottomrule
\end{tabular}%
}
}
\caption{Zero-shot image captioning on COCO test set using VLKD (RN50$\times$16), with varying number of masks.
}
\label{tab:ablation_masks}
\end{table}

\paragraph{Dataset Size of Distillation.}
In Table~\ref{tab:ablation_data}, we vary the size of dataset used for knowledge distillation. VLKD only has a slight performance drop when the size is reduced from 3M to 1M, and a sharp drop when further reduced to 100K.
\begin{table}[h]
\centering
\resizebox{\linewidth}{!}{%
\begin{tabular}{l|cc}
\toprule
 & VQAv2 (val)& COCO Caption (test) \\ \midrule \midrule
$\text{VLKD}_\text{3M}$ & 38.6 & 58.3 \\
$\text{VLKD}_\text{1M}$ & 38.3 & 56.2 \\
$\text{VLKD}_\text{100K}$ & 33.8 & 45.1 \\ \bottomrule
\end{tabular}%
}
\caption{Zero-shot performance of VLKD (ViT-B/16) on two datasets, with varying dataset size for distillation.
}
\label{tab:ablation_data}
\end{table}

\paragraph{Unfreeze CLIP Weights.} To quantitatively measure the importance of freezing the model weights of CLIP during the VLKD pre-training, we tried unfreezing CLIP's weights and conduct the VLKD pre-training using the ViT-B/16 variant on CC3M without modifying other settings. It achieves 31.7 zero-shot accuracy on the VQAv2 validation set and 44.8 CIDEr on the COCO Caption test set. We speculate that unfreezing CLIP harms its pre-trained multimodal space, which further downgrades the performance of VLKD.

\section{Conclusion} 
\label{sec:conclusion}
Recent dual-stream VLP models (e.g., CLIP) are powerful in various multimodal classification and retrieval tasks. However, their ability of multimodal  generation or pure NLP tasks is highly restricted. 
In this paper, we propose a novel knowledge distillation method to efficiently align CLIP's multimodal encoders and BART's textual encoder to the same mutlimodal space, as well as a cross-modal LM loss to consort BART encoder and decoder. This enables multimodal generation under zero-shot and also fully-finetuned settings without losing the original BART's NLP ability. Empirical results show that our model achieves new state-of-the-art zero-shot performance on VQA and excellent performance on both NLP and multimodal tasks when finetuned, demonstrating the effectiveness of our proposed method.

\clearpage
\bibliography{anthology}
\bibliographystyle{acl_natbib}

\clearpage
\appendix

\section{Hyper-parameters}
\label{sec:appendix_hyperparams}

In this section, we show the hyper-parameters of vision-language knowledge distillation (VLKD), as well as downstream task finetuning. 

For VLKD, the hyper-parameters are shown in Table~\ref{tab:appendix_kd_hyperparams}, for both two CLIP variants we explored. For finetuning multimodal downstream tasks, we use the hyper-parameters shown in Table~\ref{tab:appendix_task_hyperparams}. Within each task, we use the same setting for multiple datasets. 

For the GLUE benchmark, we use the LAMB optimizer~\citep{lamb} to train for 10 epochs. We conduct a hyper-parameter grid search with batch size=\{16, 32, 64\}, lr=\{1e-4, 5e-4, 1e-3\}, weight decay=\{1e-4, 1e-3\}. We warm up the learning rate in the first epoch, then linearly decay it to zero.

For XSUM, we directly follow the hyper-parameters used in \citet{bart}.


\begin{table}[]
\centering
\resizebox{\linewidth}{!}{%
\begin{tabular}{l|c}
\toprule
Hyper-paramters & Values \\ \midrule \midrule
Batch size & \begin{tabular}[c]{@{}c@{}}4608 (ViT-B/16 and ViT-L/14),\\ 4096 (RN50x16), 3840 (RN50x64)\end{tabular} \\
Optimizer & AdamW, $\beta=(0.99,0.999)$ \\
Learning rate & 2.4e-4 \\
Weight decay & 0.01 \\
Eps & 1e-6 \\
Temperature $\tau$ & Initialized to 0.07 \\
Warmup steps & 2\% \\
\#Epochs & 10 \\
Gradient clipping & 3.0 \\ \bottomrule
\end{tabular}%
}
\caption{Hyper-parameters of VLKD pre-training.}
\label{tab:appendix_kd_hyperparams}
\end{table}

\begin{table}[t]
\centering
\resizebox{\linewidth}{!}{%
\setlength{\tabcolsep}{4mm}{
\begin{tabular}{l|cc}
\toprule
Hyper-paramters & VQA & \begin{tabular}[c]{@{}c@{}}Image\\ captioning\end{tabular} \\ \midrule \midrule
Batch size & 72 & 64 \\
Total epochs & 10 & 10 \\
\#Masks & 2 & 6 \small{(COCO)}, 8 \small{(NoCaps)} \\
Beam search size & 1 (greedy) & 6 \\
Optimizer & \multicolumn{2}{c}{AdamW, $\beta=(0.99,0.999)$} \\
Learning rate & \multicolumn{2}{c}{1e-4} \\
Weight decay & \multicolumn{2}{c}{0.01} \\
Eps & \multicolumn{2}{c}{1e-8} \\
LR warmup & \multicolumn{2}{c}{First epoch} \\
Gradient clipping & \multicolumn{2}{c}{5.0} \\ \bottomrule
\end{tabular}%
}
}
\caption{Hyper-parameters for two multimodal tasks.}
\label{tab:appendix_task_hyperparams}
\end{table}

\begin{figure}[t!]
    \centering
    \includegraphics[width=\linewidth]{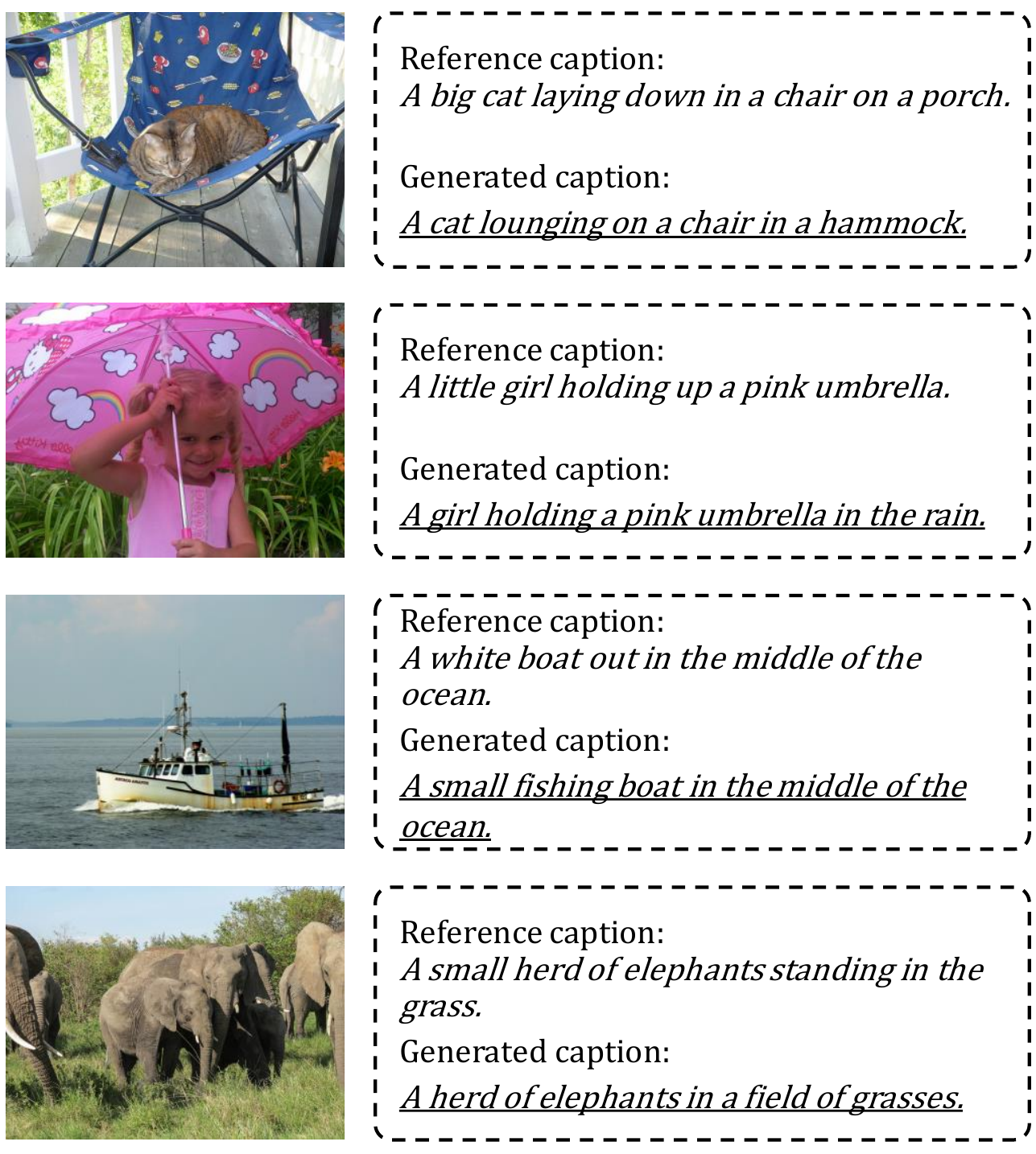}
    \caption{More examples of zero-shot image captioning.}
    \label{fig:appendix_caption}
\end{figure}

\begin{figure}[t]
    \centering
    \includegraphics[width=\linewidth]{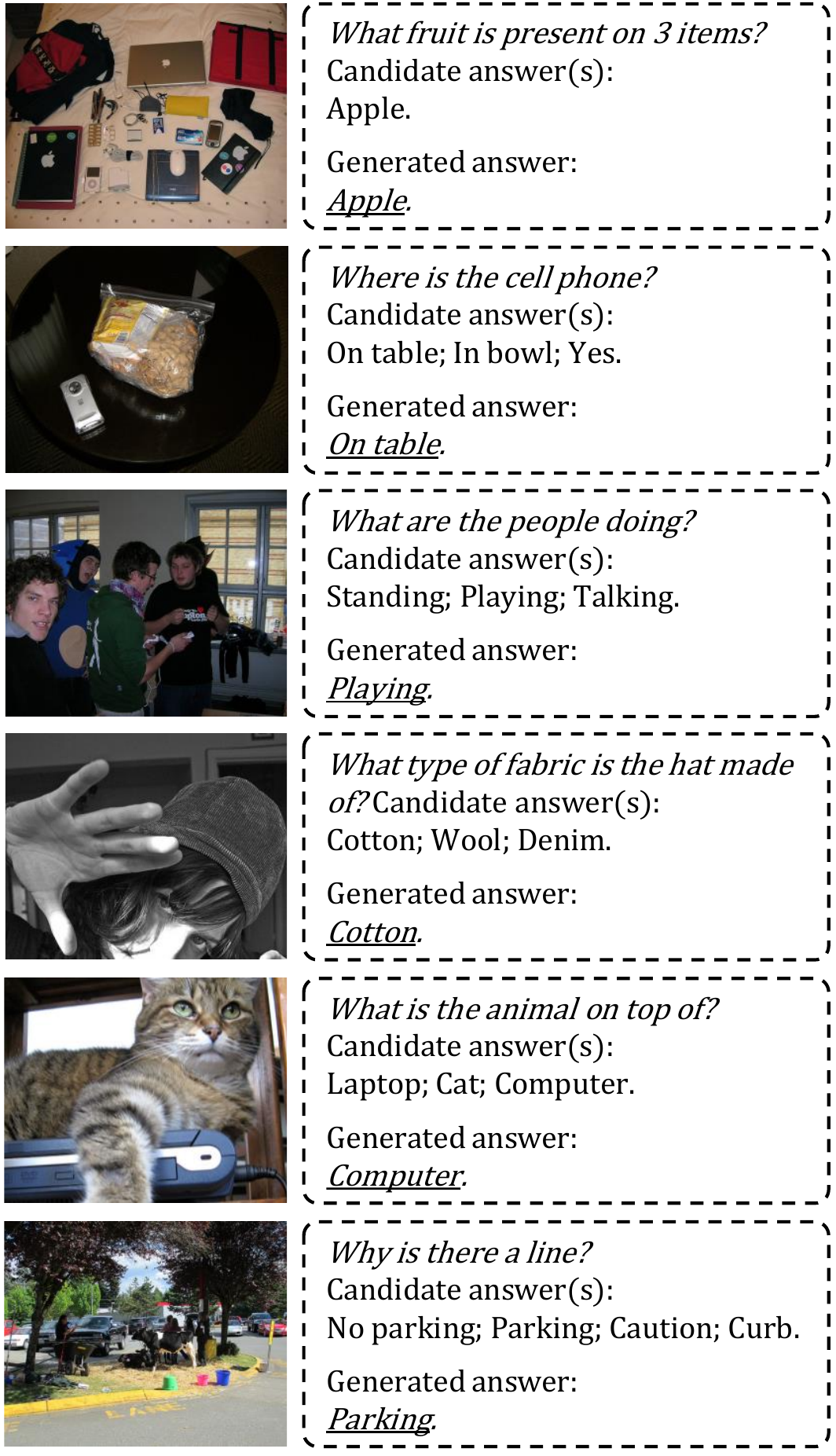}
    \caption{More examples of zero-shot VQA.}
    \label{fig:appendix_vqa}
\end{figure}

\section{More Examples of Zero-shot Inference}
In Figure~\ref{fig:appendix_caption}, we show more examples of zero-shot image captioning.
In Figure~\ref{fig:appendix_vqa}, we depict more cases of the results of zero-shot open-ended VQA.

\end{document}